\PassOptionsToPackage{usenames,dvipsnames,svgnames,table}{xcolor}
\documentclass[twoside,11pt]{article}

\usepackage[T1]{fontenc}
\usepackage[utf8]{inputenc}
\usepackage{hyperref} 
\usepackage{amsmath}
\usepackage{amssymb}
\usepackage{graphicx}
\graphicspath{{images/}}
\usepackage{subcaption}
\usepackage{booktabs}
\usepackage{transparent}
\pdfoutput=1 

\usepackage{xcolor}
\usepackage{tikz}
\usetikzlibrary{calc}
\usetikzlibrary{chains}
\usetikzlibrary{positioning,arrows}
\usetikzlibrary{decorations.pathmorphing}
\usetikzlibrary{decorations.markings}

\usepackage{listings}
\usepackage{jmlr2e}

\ShortHeadings{FastBDT}{Thomas Keck}
\firstpageno{1} 

\begin{document}

\title{FastBDT \\\vspace{6pt} \large A speed-optimized and cache-friendly implementation of stochastic gradient-boosted decision trees for multivariate classification} 
\author{\name Thomas Keck \email thomas.keck2@kit.edu \\ \addr Karlsruher Institut für Technologie, Campus Süd\\ Institut für Experimentelle Kernphysik\\ Wolfgang-Gaede-Str. 1 \\ 76131 Karlsruhe} 
\editor{}

\maketitle 

\begin{abstract}
Stochastic gradient-boosted decision trees are widely employed for multivariate classification and regression tasks.
This paper presents a speed-optimized and cache-friendly implementation for multivariate classification called FastBDT.
FastBDT is one order of magnitude faster during the fitting-phase and application-phase, in comparison 
with popular implementations in software frameworks like TMVA, scikit-learn and XGBoost.
The concepts used to optimize the execution time and performance studies are discussed in detail in this paper.
The key ideas include: 
An equal-frequency binning on the input data, which allows replacing expensive floating-point with integer operations, while
at the same time increasing the quality of the classification;
a cache-friendly linear access pattern to the input data, in contrast to usual implementations,
which exhibit a random access pattern.
FastBDT provides interfaces to C/C++, Python and TMVA. 
It is extensively used in the field of high energy physics by the Belle II experiment.
\end{abstract}


\section{Introduction}
In multivariate classification one calculates the probability of a given data-point to be signal,
characterised by a set of explanatory features $\vec{x} = \lbrace x_1, \dots, x_d \rbrace$ and a class label $y$ (signal $y = 1$ and background $y = -1$).
In supervised machine learning this involves a fitting-phase which uses training data-points with known labels
and an application-phase, during which the fitted classifier is applied to new data-points with unknown labels.
During the fitting-phase, the internal parameters (or model) of a multivariate classifier are adjusted, so that the classifier can statistically distinguish signal and background data-points. The model complexity plays an important role during the fitting-phase
and can be controlled by the hyper-parameters of the model. If the model is too simple (too complex) it will be under-fitted (over-fitted) and perform poorly on test data-points with unknown labels.
In the following the stochastic gradient-boosted decision tree algorithm \citep{Friedman2002367} and its hyper-parameters are briefly described.

\subsection{Decision Tree (DT)}
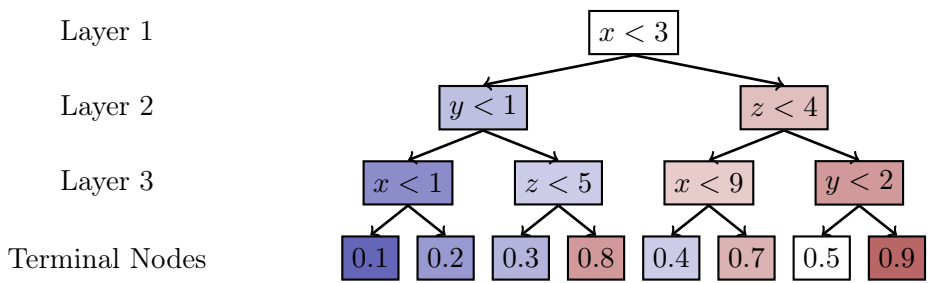
\begin{figure}
\centering
\begin{tikzpicture}[%
    box/.style={
      rectangle,
      draw=Black,
      thick,
      align=center,
      minimum height=1.5em
    },  
]
\draw (-7,0) node[align=left] {Layer 1}; 
\draw (0,0) node[box] (N1) {$x < 3$};
\draw (-7,-1) node[align=left] {Layer 2}; 
\draw (-2,-1) node[box,fill=DarkBlue!25] (N2) {$y < 1$};
\draw (2,-1) node[box,fill=DarkRed!25] (N3) {$z < 4$};
\draw[->,line width=1pt,Black] (N1.south) -> (N2.north);
\draw[->,line width=1pt,Black] (N1.south) -> (N3.north);
\draw (-7,-2) node[align=left] {Layer 3};
\draw (-3,-2) node[box,fill=DarkBlue!45] (N4) {$x < 1$};
\draw (-1,-2) node[box,fill=DarkBlue!20] (N5) {$z < 5$};
\draw (1,-2) node[box,fill=DarkRed!20] (N6) {$x < 9$};
\draw (3,-2) node[box,fill=DarkRed!40] (N7) {$y < 2$};
\draw[->,line width=1pt,Black] (N2.south) -> (N4.north);
\draw[->,line width=1pt,Black] (N2.south) -> (N5.north);
\draw[->,line width=1pt,Black] (N3.south) -> (N6.north);
\draw[->,line width=1pt,Black] (N3.south) -> (N7.north);
\draw (-7,-3) node[align=left] {Terminal Nodes};
\draw (-3.5,-3) node[box,fill=DarkBlue!60] (N8) {$0.1$};
\draw (-2.5,-3) node[box,fill=DarkBlue!40] (N9) {$0.2$};
\draw (-1.5,-3) node[box,fill=DarkBlue!30] (N10) {$0.3$};
\draw (-0.5,-3) node[box,fill=DarkRed!40] (N11) {$0.8$};
\draw (0.5,-3) node[box,fill=DarkBlue!20] (N12) {$0.4$};
\draw (1.5,-3) node[box,fill=DarkRed!30] (N13) {$0.7$};
\draw (2.5,-3) node[box,fill=DarkRed!0] (N14) {$0.5$};
\draw (3.5,-3) node[box,fill=DarkRed!60] (N15) {$0.9$};
\draw[->,line width=1pt,Black] (N4.south) -> (N8.north);
\draw[->,line width=1pt,Black] (N4.south) -> (N9.north);
\draw[->,line width=1pt,Black] (N5.south) -> (N10.north);
\draw[->,line width=1pt,Black] (N5.south) -> (N11.north);
\draw[->,line width=1pt,Black] (N6.south) -> (N12.north);
\draw[->,line width=1pt,Black] (N6.south) -> (N13.north);
\draw[->,line width=1pt,Black] (N7.south) -> (N14.north);
\draw[->,line width=1pt,Black] (N7.south) -> (N15.north);
\end{tikzpicture}
\caption{Three layer DT: A given test data-point (with unknown label) traverses the tree from top to bottom. At each node of the tree a binary decision is made until a terminal node is reached. The probability of the test data-point to be signal
is the signal-fraction (number stated in terminal node layer) of all training data-points (with known label), which ended up in the same terminal node.}
\label{fig:DecisionTree}
\end{figure}

A DT performs a classification using a number of consecutive cuts (see Figure\ref{fig:DecisionTree}).
The maximum number of consecutive cuts is a hyper-parameter and is called the \texttt{depth of the tree $D$}.

The cuts are determined during the fitting-phase using a training sample with known labels. 
At each node only training data-points which passed the preceding cuts are considered. 
For each feature at each node a cumulative probability histogram (CPH) for signal and background is calculated, respectively.
The histograms are used to determine the separation gain for a cut at each position in these histograms.
The feature and cut-position (or equivalently bin) with the highest separation gain are used as the cut for the node.
Hence each cut locally maximises the separation gain between signal and background on the given training sample.

The predictions of a deep DT is often dominated by statistical fluctuations in the training data-points.
In consequence, the classifier is over-fitted and performs poorly on new data-points.
There are pruning algorithms which automatically remove cuts prone to over-fitting from the DT \citep{Pruning}. These algorithms are not further discussed here.
A detailed description of decision trees is available in \citet{cart84}.

\subsection{Boosted Decision Tree (BDT)}
A BDT constructs a more robust classification model by sequentially constructing shallow DTs during the fitting-phase.
The DTs are fitted so that the expectation value of a negative binomial log-likelihood loss-function is minimized.
The depth of the individual DT is strongly limited to avoid over-fitting.
Therefore a single DT separates signal and background only roughly and is a so-called weak-learner\footnote{A simple model with few parameters.}.
By using many weak-learners a well-regularized classifier with large separation power is constructed. 
The \texttt{number of trees $N$} (or equivalently the number of boosting steps) and the \texttt{learning rate $\eta$} are additional hyper-parameters of this model. The two parameters are anti-correlated meaning decreasing the value of $\eta$ increases the best value for $N$.

\subsection{Gradient Boosted Decision Tree (GBDT)}
A GBDT uses gradient-decent in each boosting step to re-weight the original training sample.
In consequence, data-points which are hard to classify (often located near the optimal separation hyper-plane) gain influence during the training.
A boost-weight calculated for each terminal node during the fitting-phase is used as output of each DT instead of the signal-fraction.
The probability of a test data-point (with unknown label) to be signal is the sigmoid-transformed sum of the outputted boost-weights of each tree.
The complete algorithm is derived and discussed in detail by \citet{Friedman00greedyfunction}.

\subsection{Stochastic Gradient Boosted Decision Tree (SGBDT)}
A SGBDT uses a randomly drawn (without replacement) sub-sample instead of the full training sample in each boosting step during the fitting phase.
This approach further increases the robustness against over-fitting, because the statistical fluctuations in the training sample are averaged out in the sum over all trees.
The incorporation of randomization into the procedure was extensively studied by \citet{Friedman2002367}.
The fraction of samples used in each boosting step is another hyper-parameter called the \texttt{sub-sampling rate $\alpha$}.

\subsection{Related Work}
In general there are two approaches to increase the execution speed of an algorithm: Modify the algorithm itself
or optimize its implementation. 

It is easy to see that the first approach has large potential and there are several authors
which investigated this approach for the SGBDT algorithm: Already the original paper \citep{Friedman00greedyfunction} on GBDTs showed that $90$ \% to $95$ \% of the training
data-points can be removed from the fitting-phase after some boosting steps without sacrificing accuracy of the classifier.
Another approach was presented in \citet{icml2013_appel13} where a subset of the training data was used to prune underachieving features early during
the fitting phase without affecting the final performance.
Traditional boosting as discussed above treats the tree learner as a black box, however it is possible to
exploit the underlying tree structure to achieve higher accuracy and smaller (hence faster) models \citep{johnson2014}.

FastBDT uses a complementary approach and gains an order of magnitude in execution time by optimizing mainly the implementation of the algorithm.
The techniques mentioned above could be implemented in order to get even higher speed-ups.

\section{FastBDT implementation}
On modern hardware it is difficult to predict the execution time, e.g. in terms of spent CPU cycles, because 
there are many mechanisms built into modern CPUs to exploit parallelisable code execution and memory access patterns.
In consequence it is important to benchmark all performance optimisations. In this work perf \citep{PerfTool}, valgrind \citep{Nethercote:2007:VFH:1273442.1250746} and
std::chrono::high\_resolution\_clock \citep{ISO:2012:III} were used to benchmark the execution time and identify critical code sections.
The most time consuming code section in the SGBDT algorithm is the calculation of the cumulative probability histograms (CPH),
which are required in order to calculate the best-cut at each node of the tree.
The main concepts used in the implementation of the fitting-phase of FastBDT are described in the following.

\begin{figure}
\centering
\begin{tikzpicture}[%
    box/.style={
      rectangle,
      draw=Black,
      thick,
      align=left,
      minimum height=1.5em
    },  
]
\draw (-4,0) node[align=left] {\texttt{struct of arrays}}; 
\draw (0.0,0) node[box] {$\: \textcolor{DarkBlue}{x_1}\ \mathbf{\textcolor{DarkRed}{x_2}}\ \mathbf{\textcolor{DarkRed}{x_3}}\ \textcolor{DarkBlue}{x_4}\ \mathbf{\textcolor{DarkRed}{x_5}}\ \textcolor{DarkBlue}{x_6}\:$};
\draw (3.5,0) node[box] {$\: \textcolor{DarkBlue}{y_1}\ \mathbf{\textcolor{DarkRed}{y_2}}\ \mathbf{\textcolor{DarkRed}{y_3}}\ \textcolor{DarkBlue}{y_4}\ \mathbf{\textcolor{DarkRed}{y_5}}\ \textcolor{DarkBlue}{y_6}\:$};
\draw (6.9,0) node[box] {$\: \textcolor{DarkBlue}{z_1}\ \mathbf{\textcolor{DarkRed}{z_2}}\ \mathbf{\textcolor{DarkRed}{z_3}}\ \textcolor{DarkBlue}{z_4}\ \mathbf{\textcolor{DarkRed}{z_5}}\ \textcolor{DarkBlue}{z_6}\:$};
\draw (-4,0.8) node[align=left] {\texttt{array of structs}}; 
\draw (-0.95,0.8) node[box] {$\textcolor{DarkBlue}{x_1}\ \textcolor{DarkBlue}{y_1}\ \textcolor{DarkBlue}{z_1}$};
\draw (0.78,0.8) node[box] {$\mathbf{\textcolor{DarkRed}{x_2}}\ \mathbf{\textcolor{DarkRed}{y_2}}\ \mathbf{\textcolor{DarkRed}{z_2}}$};
\draw (2.58,0.8) node[box] {$\mathbf{\textcolor{DarkRed}{x_3}}\ \mathbf{\textcolor{DarkRed}{y_3}}\ \mathbf{\textcolor{DarkRed}{z_3}}$};
\draw (4.3,0.8) node[box] {$\textcolor{DarkBlue}{x_4}\ \textcolor{DarkBlue}{y_4}\ \textcolor{DarkBlue}{z_4}$};
\draw (6.05,0.8) node[box] {$\mathbf{\textcolor{DarkRed}{x_5}}\ \mathbf{\textcolor{DarkRed}{y_5}}\ \mathbf{\textcolor{DarkRed}{z_5}}$};
\draw (7.8,0.8) node[box] {$\textcolor{DarkBlue}{x_6}\ \textcolor{DarkBlue}{y_6}\ \textcolor{DarkBlue}{z_6}$};
\end{tikzpicture}
\caption{Possible memory layouts of the input features. Shown are three features $x$, $y$ and $z$ and six data-points.
The bold-red and blue colouring will take on different meanings as we go through the concepts described below.}
\label{fig:MemoryLayouts}
\end{figure}
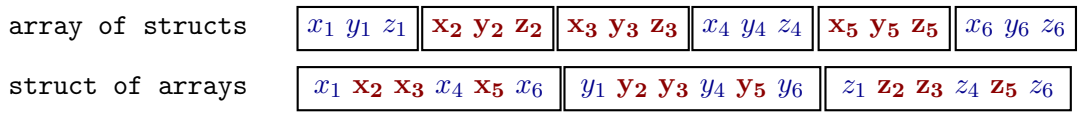

\subsection{Memory Access Patterns}
At first we consider the memory layout of the training data. Each training data-point consists of $d$ continuous features, one boolean label (signal or background) and an optional continuous weight. In total there are $M$ training data-points. There are two commonly used memory layouts in this situation: \texttt{array of structs} and \texttt{struct of arrays} (see Figure \ref{fig:MemoryLayouts}).

CPU caches assume \emph{spatial locality} \citep{Kowarschik2003}, which means that if the values in memory are accessed in linear order there is a high probability that 
they are already cached. In addition, CPU caches assume \emph{temporal locality} \citep{Kowarschik2003}, which means that frequently accessed values in memory are cached as well.
Finally, the CPU assumes \emph{branch locality}, meaning only a few conditional jumps occure and therefore instructions can be efficiently prefetched and pipelined.
The key idea is to use spatial locality (i.e., a linear memory access pattern) for the caching of the large feature input data, temporal locality for the caching of the small cumulative probability histograms (CPH), while at the same time avoiding conditional jumps as much as possible to ensure branch locality.

FastBDT uses an \texttt{array of structs} memory layout and determines the cuts for all features and all nodes in the same layer of the tree simultaneously.
Several sources of conditional jumps and non-uniform memory access are considered:

\paragraph{Signal and Background}
The CPHs are calculated separately for signal and background (imagine the red data-points are signal and the blue data-points are background in Figure \ref{fig:MemoryLayouts}).
This leads to additional conditional jumps in both memory layouts.
Hence FastBDT stores and processes signal and background data-points separately.
Additionally this reduces the amount of CPH data which has to be cached by a factor of two, and saves one conditional jump during the filling of the CPHs.

\paragraph{Multiple Nodes per Layer}
In each layer of the decision tree the algorithm has to calculate the optimal cuts for multiple nodes, where each node is optimized with respect to a distinct subset of the training data-points (imagine the blue data-points belong to node A and the red data-points belong to node B in Figure \ref{fig:MemoryLayouts}).
In both memory layouts consecutive memory access is only possible if the CPHs are calculated for all nodes in the current layer in parallel.
Therefore FastBDT calculates the CPHs in different nodes at the same time (the CPH data is likely to be cached due to temporal locality), which allows accessing the values in direct succession (hence the input data is likely to be cached due to spatial locality).
In contrast, other popular implementations determine the cuts node-after-node. Hence the data-points which have to be accessed at the current node depend on the preceding nodes and therefore this commonly used approach exhibits random jumps during the memory access regardless of the memory layout.

\paragraph*{Stochastic sub-sampling}
Only a fraction of the training data-points is used during the fitting of each tree (imagine the blue data-points are used and the red data-points are disabled in Figure \ref{fig:MemoryLayouts}).
The \texttt{array of structs} allows consecutive memory access within a data-point without conditional jumps.
As a consequence FastBDT uses this memory layout, but has to calculate all the CPHs for the features in parallel as well.
In the \texttt{struct of arrays} layout the conditional jumps cannot be avoided without re-arranging the data in the memory.

\subsection{Preprocessing}
Continuous input features are represented by floating point numbers.
However DTs only use the order statistic of the features (in contrast to, e.g., artificial neural networks).
Hence the algorithm only compares the values to one another and doesn't use the values themselves.
Therefore FastBDT performs an equal-frequency binning on the input features and maps them to integers.
This has several advantages: Usually integer operations can be performed faster by the CPU; the integers can
directly be used as indices for the CPHs during the calculation of the best-cuts; and the quality of the separation
is often improved because the shape of the input feature distribution (which may contain sharp peaks or heavy tails) are mapped to a uniform distribution.

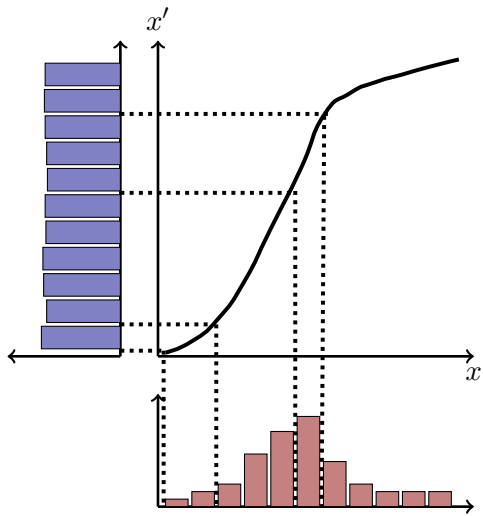
\begin{figure}
\begin{center}
\begin{tikzpicture}[%
    box/.style={
      rectangle,
      draw=Black,
      thick,
      align=left,
      minimum height=1.5em
    },  
]
\draw[->,line width=1pt,Black] (0,0) -> (0,4.2) node[above] {$x'$} ;
\draw[->,line width=1pt,Black] (0,0) -> (4.2,0) node[below] {$x$} ;

\draw[->,line width=1pt,Black] (-0.5,0) -> (-2, 0) ;
\draw[->,line width=1pt,Black] (-0.5,0) -> (-0.5,4.2);

\draw[->,line width=1pt,Black] (0,-2) -> (0,-0.5);
\draw[->,line width=1pt,Black] (0,-2) -> (4.2,-2);

\draw [fill=DarkRed!50] (0.1,-2) rectangle (0.4,-1.9);
\draw [fill=DarkRed!50] (0.45,-2) rectangle (0.75,-1.8);
\draw [fill=DarkRed!50] (0.8,-2) rectangle (1.1,-1.7);
\draw [fill=DarkRed!50] (1.15,-2) rectangle (1.45,-1.3);
\draw [fill=DarkRed!50] (1.5,-2) rectangle (1.8,-1.0);
\draw [fill=DarkRed!50] (1.85,-2) rectangle (2.15,-0.8);
\draw [fill=DarkRed!50] (2.2,-2) rectangle (2.5,-1.4);
\draw [fill=DarkRed!50] (2.55,-2) rectangle (2.85,-1.7);
\draw [fill=DarkRed!50] (2.9,-2) rectangle (3.2,-1.8);
\draw [fill=DarkRed!50] (3.25,-2) rectangle (3.55,-1.8);
\draw [fill=DarkRed!50] (3.6,-2) rectangle (3.9,-1.8);

\draw [fill=DarkBlue!50] (-1.55,0.1) rectangle (-0.5, 0.4);
\draw [fill=DarkBlue!50] (-1.48,0.45) rectangle (-0.5,0.75);
\draw [fill=DarkBlue!50] (-1.52,0.8) rectangle (-0.5, 1.1);
\draw [fill=DarkBlue!50] (-1.53,1.15) rectangle (-0.5,1.45);
\draw [fill=DarkBlue!50] (-1.49,1.5) rectangle (-0.5,1.8);
\draw [fill=DarkBlue!50] (-1.5,1.85) rectangle (-0.5,2.15);
\draw [fill=DarkBlue!50] (-1.47,2.2) rectangle (-0.5,2.5);
\draw [fill=DarkBlue!50] (-1.48,2.55) rectangle (-0.5,2.85);
\draw [fill=DarkBlue!50] (-1.5,2.9) rectangle (-0.5,3.2);
\draw [fill=DarkBlue!50] (-1.51,3.25) rectangle (-0.5,3.55);
\draw [fill=DarkBlue!50] (-1.5,3.6) rectangle (-0.5,3.9);

\draw[-,dotted,line width=1.5pt,Black] (-0.5,0.075) -> (0.1,0.075);
\draw[-,dotted,line width=1.5pt,Black] (-0.5,0.425) -> (0.7,0.425);
\draw[-,dotted,line width=1.5pt,Black] (0.075, 0.075) -> (0.075,-2);
\draw[-,dotted,line width=1.5pt,Black] (0.775, 0.425) -> (0.775,-2);

\draw[-,dotted,line width=1.5pt,Black] (-0.5,2.175) -> (1.8,2.175);
\draw[-,dotted,line width=1.5pt,Black] (-0.5,3.225) -> (2.3,3.225);
\draw[-,dotted,line width=1.5pt,Black] (1.825, 2.2) -> (1.825,-2);
\draw[-,dotted,line width=1.5pt,Black] (2.21, 3.2) -> (2.175,-2);

\draw [black, line width=1.5pt] plot [smooth, tension=1] coordinates { (0.1,0.05) (0.45, 0.2) (0.8,0.5) (1.15,1.0) (1.5,1.7) (1.9, 2.5) (2.2, 3.2) (2.55, 3.5) (2.9, 3.65) (3.25, 3.75) (3.6, 3.85) (4.0,3.95)};
\end{tikzpicture}
\end{center}
\caption{Equal-frequency binning. The feature $x$ is binned, so that in each bin there is roughly the same number of data-points. The bin boundaries are indicated by the dotted lines.}
\end{figure}

This preprocessing is only done during the fitting-phase. Once all cuts are determined the inverse transformation is used
to map the integers used in the cuts back to the original floating point numbers. Therefore there is no runtime-overhead 
during the application-phase due to this preprocessing.

\subsection{Parallelism}
Usually one can distinguish between two types of parallelism: task-level parallelism (multiple processors, multiple cores per processor, multiple threads
per core) and instruction-level parallelism (instruction pipelining, multiple execution units and ports, vectorization).
It is possible to use task-level parallelism to reduce the execution time during the fitting-phase of the SGBDT algorithm. However FastBDT was not designed to do so,
since our use-cases typically require fitting many classifiers in parallel and therefore already exploit task-level parallelism effectively, or use a shared infrastructure, where we are only interested in minimizing the total CPU time.
Other implementations like XGBoost do take advantage of this type of parallelism.

The application-phase is embarrassingly parallel and task-level parallelism can be used
by all implementations even if not directly supported\footnote{E.g. by running the loop over the data-points in parallel using OpenMPI \citep{openmp15} \texttt{\#pragma omp parallel for} like it is done in XGBoost after calling its predict function.}.

On the other hand instruction-level parallelism is difficult to exploit in the SGBDT algorithm (in the fitting-phase and the application-phase) due to the large number of conditional jumps intrinsic to the algorithm, which strongly limits the use of instruction pipelining and vectorization.
However, due to the chosen memory layout and preprocessing many conditional jumps in the algorithm can be replaced by index operations.
Figure \ref{lst:branchlocality} show an example for such an optimization (branch locality)

\begin{figure}
\begin{subfigure}[b]{0.49\textwidth}
\centering
\begin{lstlisting}[language=C++,numbers=none,xleftmargin=0mm]
 int a = 0;
 int b = 0;
 for(int i=0; i<1e9; ++i) {
  if(rand()%2 == 0) a++;
  else              b++;
 }
 cout<<a<<" "<<b<<endl;
\end{lstlisting}
\caption{Straight-forward implementation --\\ Execution time 10.1 $\mathrm{sec}$}
\end{subfigure}
\begin{subfigure}[b]{0.50\textwidth}
\centering
\begin{lstlisting}[language=C++,numbers=none,xleftmargin=0mm]
int a[] = {0,0};

for(int i=0; i<1e9; ++i) {
  a[rand()%2]++;

} 
cout<<a[0]<<" "<<a[1]<<endl;
\end{lstlisting}
\caption{If statement replaced by array lookup -- Execution time 6.9 $\mathrm{sec}$}
\end{subfigure}
\caption{Branch locality optimisation example which increments two counters randomly.
Both codes were compiled with the highest optimization level (-Ofast) available using g++ 4.8.4. The optimised version
is $30$\% faster.}
\label{lst:branchlocality}
\end{figure}

\section{Comparison}

FastBDT (development version 24.04.2016)\footnote{git hash ce39fa9ac8cd0e94a5b7d5cdef34300e5d372a63} was compared against other SGBDT implementations used  usually in high energy physics:
\begin{itemize}
 \setlength\itemsep{0.0em}
\item TMVA \citep{Hocker:2007ht} (ROOT version 6.06/00) -- The multivariate analysis package of the ROOT \citep{Brun199781} data analysis library developed by CERN (GPLv2 license);
\item scikit-learn \citep{scikit-learn} (version 0.17.1) -- A machine learning library written in python (BSD license);
\item xgboost \citep{XGBoost} (development version 22.04.2016)\footnote{git hash b3c9e6a0db0a7eb755949ac6b26e3ef805738350} -- A modern implementation of BDTs which performed well in the Higgs Boson challenge \citep{xgboostHiggsChallenge} (Apache license).
\end{itemize}
The used dataset contains 1 million data-points, 35 features and a binary target.
It was split into equal parts, used during the fitting and application phase, respectively.
The dataset was produced using Monte Carlo simulation of the decay of a $D^0$ meson into one charged pion, one neutral pion and one kaon.
A common classification problem in high energy physics. However, the nature of the data has no influence on the execution time of the SGBDT algorithm
in the considered implementations, since no optimizations which prune features using their estimated separation power, as described in \citet{icml2013_appel13}, are employed.

In a first \textbf{preprocessing} step the dataset is converted into the preferred data-format of each implementation:
\begin{itemize}
 \setlength\itemsep{0.0em}
\item ROOT \texttt{TTree} for TMVA,
\item numpy \texttt{ndarray} for scikit-learn,
\item \texttt{DMatrix} for XGBoost
\item and \texttt{FastBDT::EventSample} for FastBDT.
\end{itemize}
Afterwards, the \textbf{fitting} and \textbf{application} steps are performed for each implementation.
Each step is measured individually using std::chrono::high\_resolution\_clock. The preprocessing time is small\footnote{less than $< 20 \%$ of the training time} with respect to the fitting time
for most ($> 95 \%$) of the investigated hyper-parameter configurations. All results stated below are valid as well if one considers the preprocessing 
as part of the fitting-phase.

The execution time of each implementation is measured five times for each considered set of hyper-parameters.
If not stated otherwise the following hyper-parameters are chosen:
\begin{itemize}
 \setlength\itemsep{0.0em}
\item \texttt{depth of the trees} $= 3$
\item \texttt{number of trees} $= 100$
\item \texttt{number of features} $= 35$
\item \texttt{number of training data-points} $= 500000$
\item \texttt{sampling-rate} $= 0.5$
\item \texttt{shrinkage} $= 0.1$
\end{itemize}
All implementations have additional hyper-parameters which are not shared by the other implementations.
The respective default values are used in these cases.

Two versions of XGBoost are considered: The single-core (named just XGBoost in the following) and multi-core version (named XGBoost-i7 in the following).
In addition a simple multi-core version of FastBDT is considered in the application-phase measurements (named FastBDT-i7 in the following).

All measurements are performed on an Intel(R) Core(TM) i7-4770 CPU (@ 3.40GHz) with a main memory of $32\ \mathrm{GigaByte}$.
The code used to perform the measurements can be found in the FastBDT repository.

\subsection{Fitting-Phase}
\begin{figure}
\centering
\begin{subfigure}[b]{0.49\textwidth}
\includegraphics[width=1.1\textwidth]{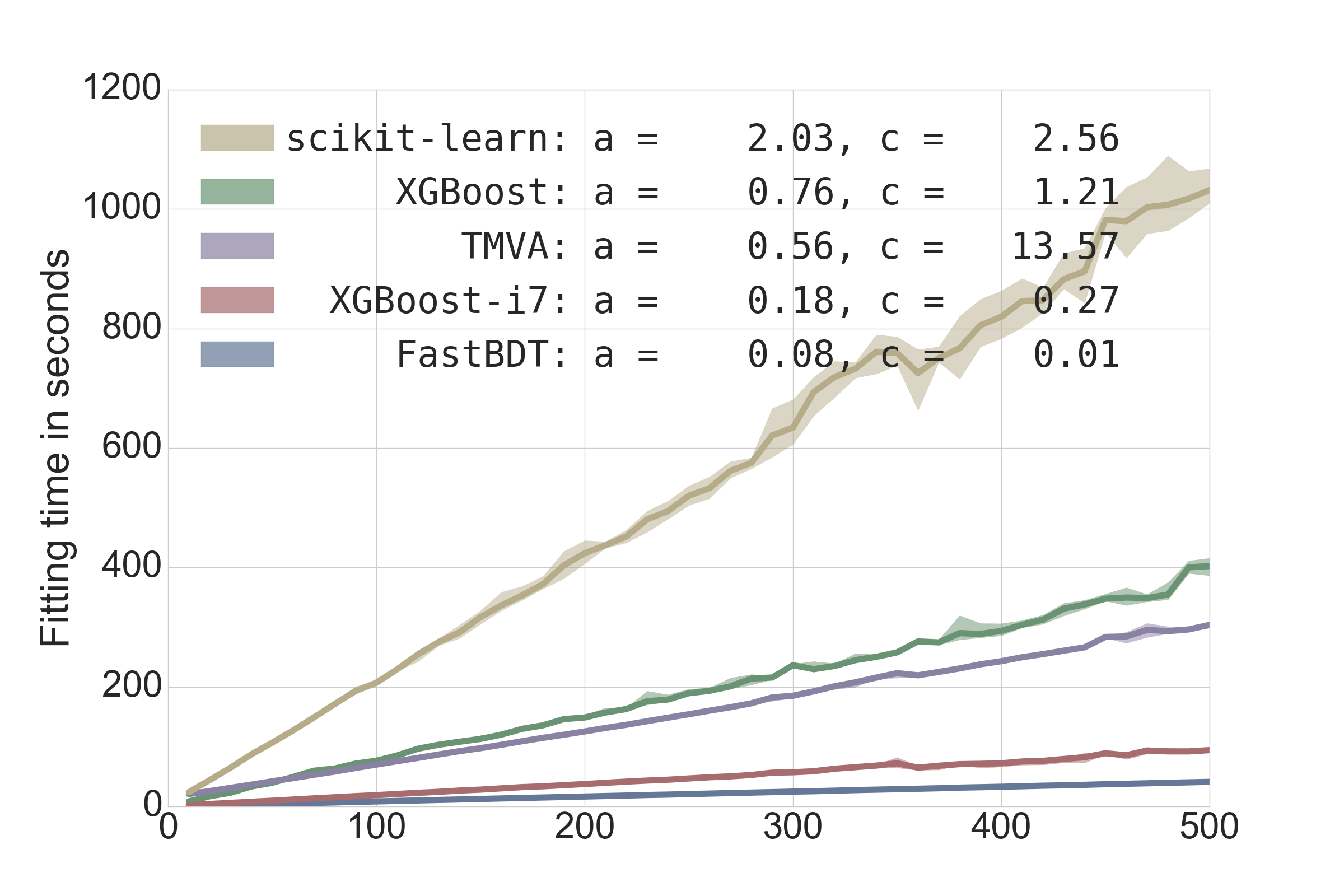}
\caption{\texttt{number of trees}}
\label{fig:FitnTrees}
\end{subfigure}\hfill
\begin{subfigure}[b]{0.49\textwidth}
\includegraphics[width=1.1\textwidth]{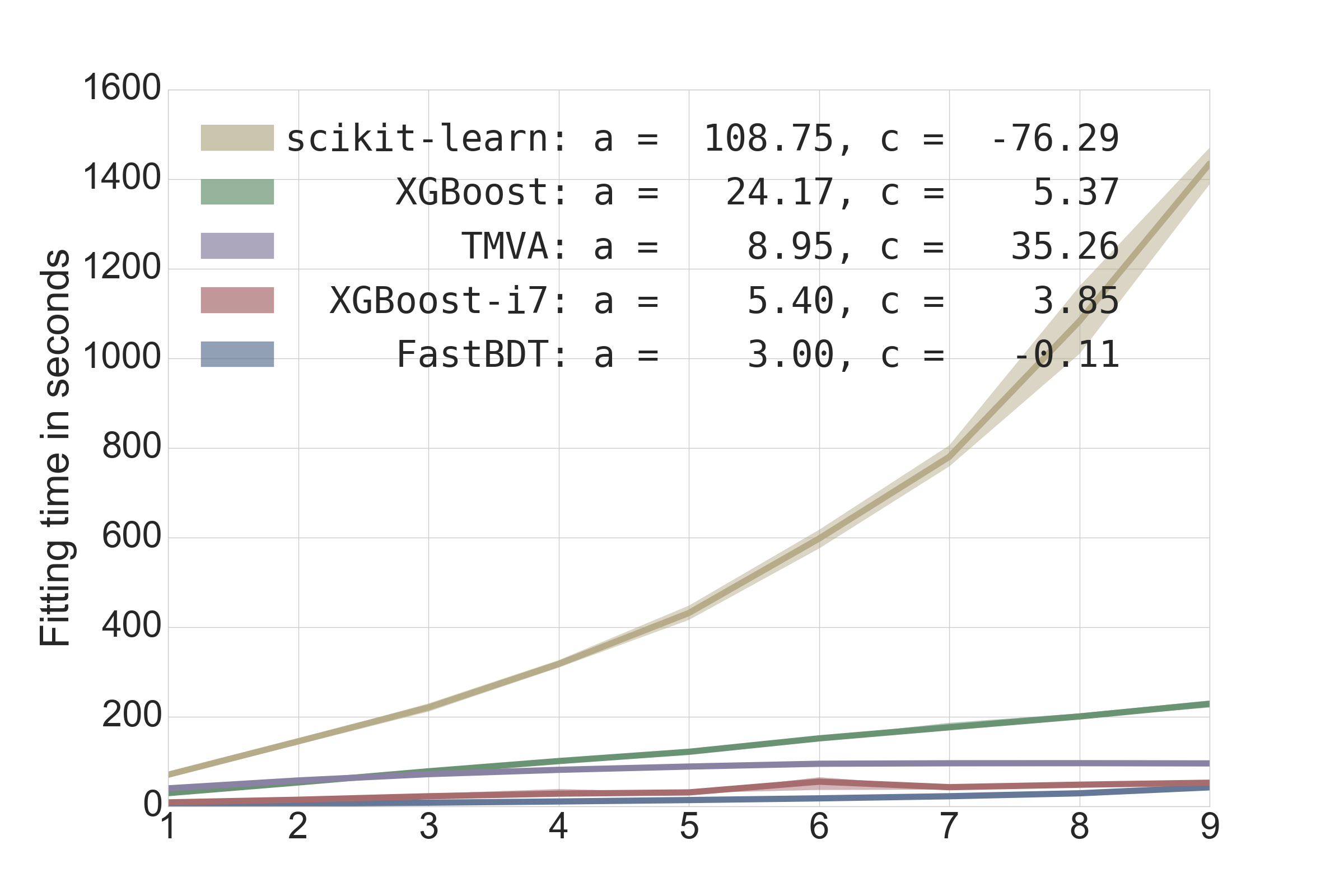}
\caption{\texttt{depth of the trees}}
\label{fig:Fitdepth}
\end{subfigure}\hfill
\begin{subfigure}[b]{0.49\textwidth}
\includegraphics[width=1.1\textwidth]{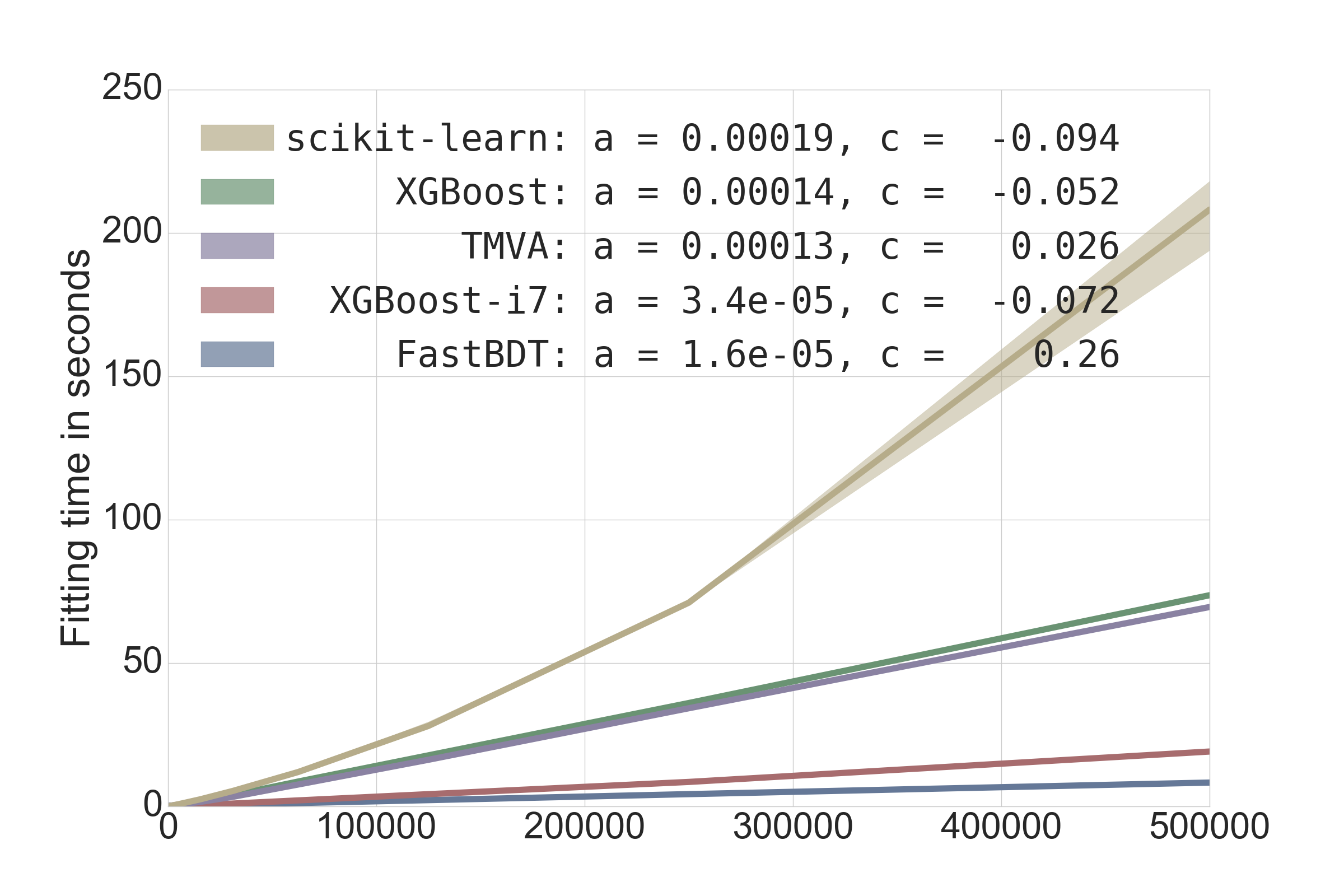}
\caption{\texttt{number of training data-points}}
\label{fig:FitnEvents}
\end{subfigure}\hfill
\begin{subfigure}[b]{0.49\textwidth}
\includegraphics[width=1.1\textwidth]{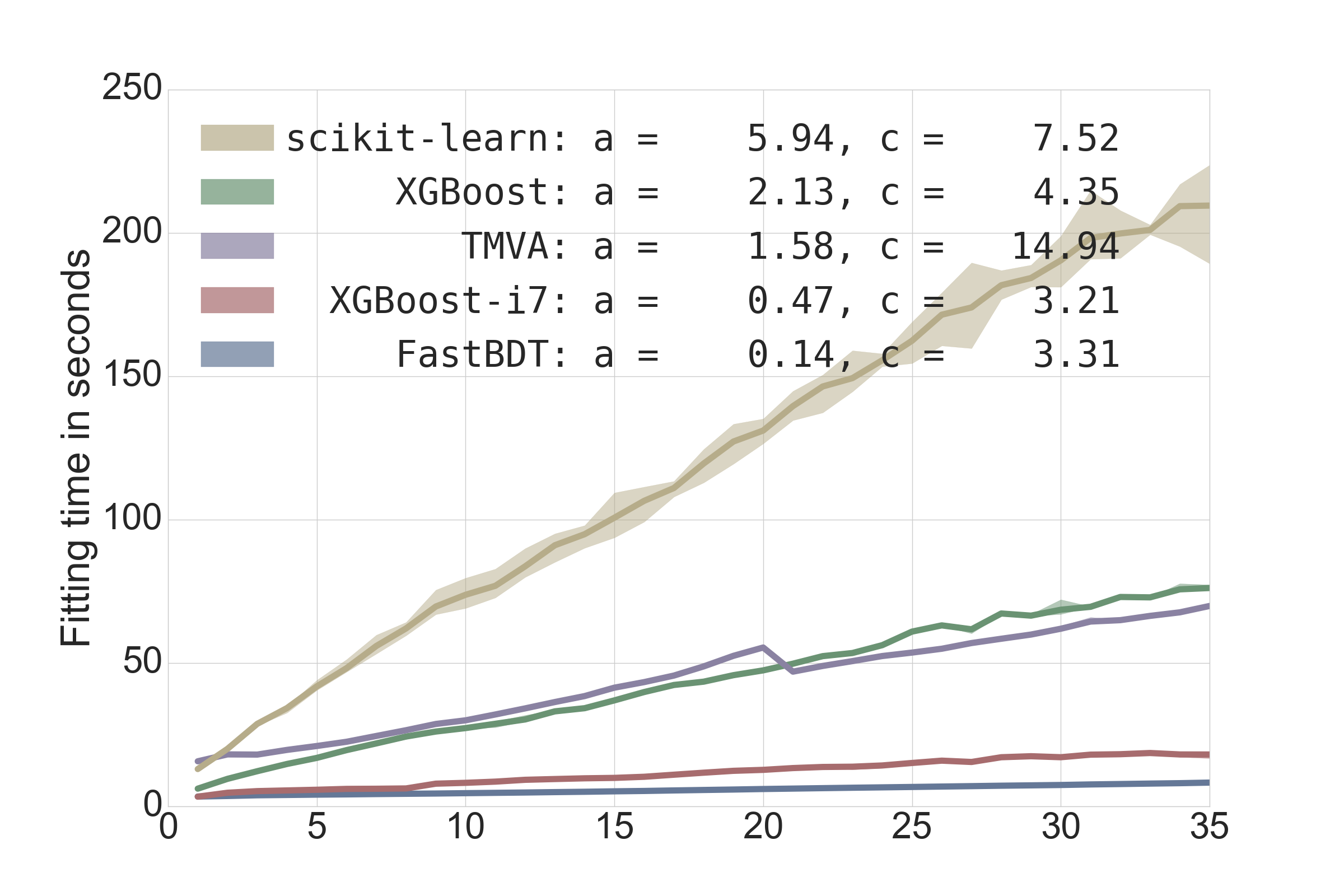}
\caption{\texttt{number of features}}
\label{fig:FitnFeatures}
\end{subfigure}
\begin{subfigure}[b]{0.49\textwidth}
\includegraphics[width=1.1\textwidth]{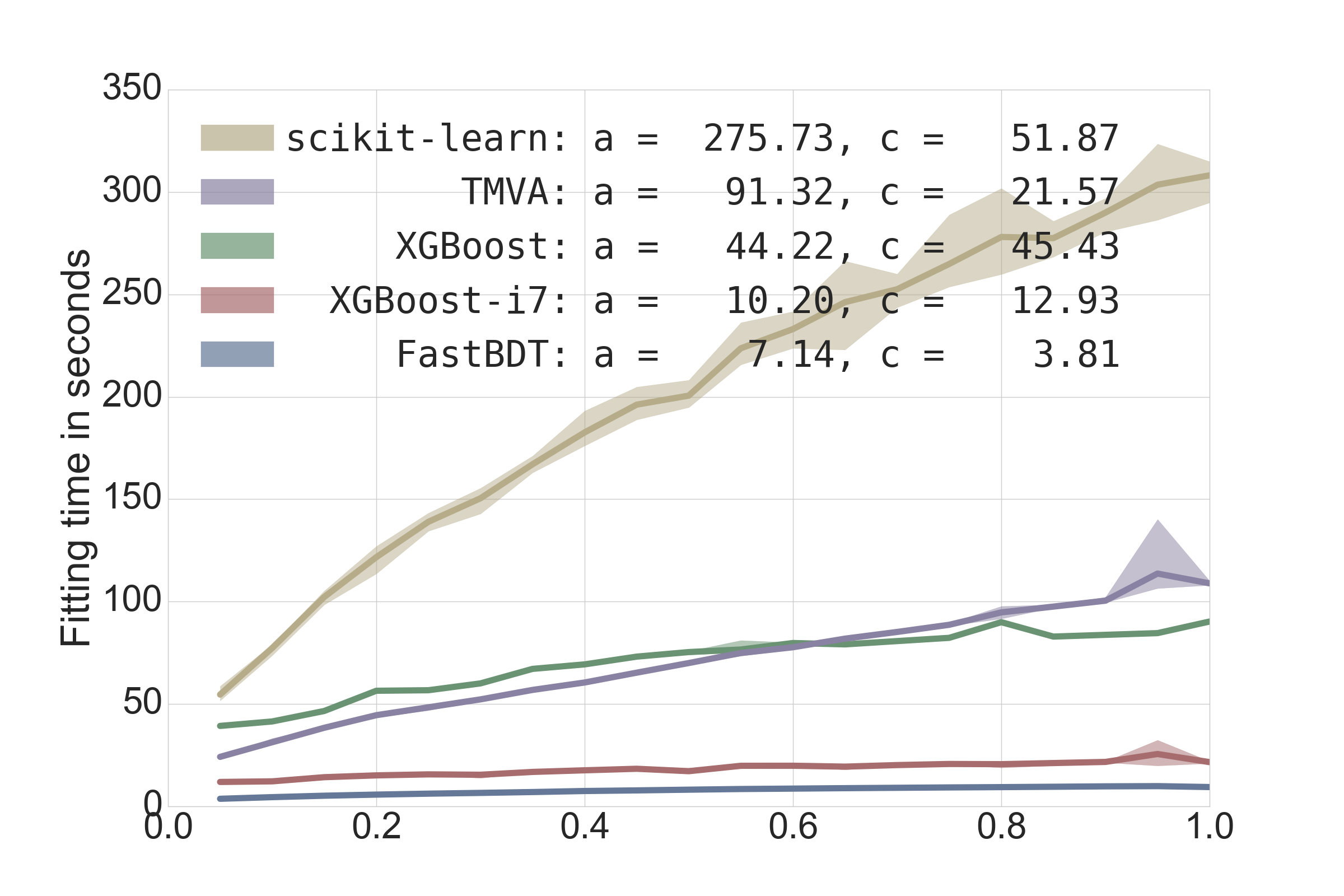}
\caption{\texttt{sampling rate}}
\label{fig:FitSamplingRate}
\end{subfigure}
\caption{Runtime behaviour during the fitting-phase of different hyper-parameters. The $y$ axes show the training time. The $x$ axes show the varied hyper-parameter. Each configuration was measured five times:
The mean values are shown as solid lines, whereas the minimal (maximal) values define the lower (upper) bound of the shaded band around the lines. A linear regression $y = a \cdot x + c$ was performed
using the minimal values of each configuration. The order of the legend entries is the same as the position of the solid lines at the right limit of each plot.}
\label{fig:TrainingTime}
\end{figure}

In general it is expected that the fitting-phase runtime of the algorithms scale linearly in \texttt{depth of the trees}, \texttt{number of trees},
\texttt{number of features}, \texttt{number of training data-points} and \texttt{sampling-rate}.
As can be seen from Figure \ref{fig:TrainingTime} these expectations are not always fulfilled.
For each varied hyper-parameter the gradient $a$ and offset $c$ was fitted using an ordinary linear regression.

FastBDT outperforms all other implementations during the fitting phase, including the multi-core version of XGBoost.
scikit-learn is the slowest contestant and also violates the expected linear runtime behaviour for the  \texttt{depth of the trees} (see \ref{fig:Fitdepth}) and the \texttt{number of training data-points}  (see \ref{fig:FitnEvents}). It is not clear why this is the case.
TMVA shows a constant runtime for  \texttt{depth of the trees} $> 6$  (see \ref{fig:Fitdepth}); this can be explained by the default limit on the minimum number of data-points per node of $5 \% $.
Hence TMVA stopped growing the trees deeper than six layers. The linear increase in the runtime for the \texttt{number of features} has a constant drop at $20$, where TMVA stops filling pairwise histograms
for all features, which are used for evaluation and plotting purposes only.
FastBDT and XGBoost have a linear runtime behaviour in all considered hyper-parameters, as expected.
The multi-core version of XGBoost has an average speedup of $3.83$ using 8 cores (including hyper-threading)
over the single-core variant and shows a better performance for deep trees.
Detailed speedup comparisons between all tested implementations are stated in Table \ref{tab:fittingphase}.

\begin{table}
\centering
\setlength\tabcolsep{2pt}
\caption{Average ratio of fitting-phase runtime over all considered hyper-parameter configurations.}
\label{tab:fittingphase}
\begin{tabular}{c|rrrrr}
	\toprule
	           & FastBDT & scikit-learn & XGBoost & XGBoost-i7 &  TMVA \\ \midrule
	 FastBDT   &         & $21.6$  & $7.8$  & $2.0$ & $7.3$ \\
	 scikit-learn   &  $0.1$  &         & $0.4$  & $0.1$ & $0.4$ \\
	 XGBoost   &  $0.2$  & $2.8$   &        & $0.3$ & $1.0$ \\
	XGBoost-i7 &  $0.6$  & $10.7$  & $3.8$  &       & $3.7$ \\
	   TMVA    &  $0.2$  & $3.2$   & $1.1$  & $0.3$ &   \\ \bottomrule
\end{tabular}
\end{table}

\subsection{Application-Phase}
\begin{figure}
\centering
\begin{subfigure}[b]{0.49\textwidth}
\includegraphics[width=1.1\textwidth]{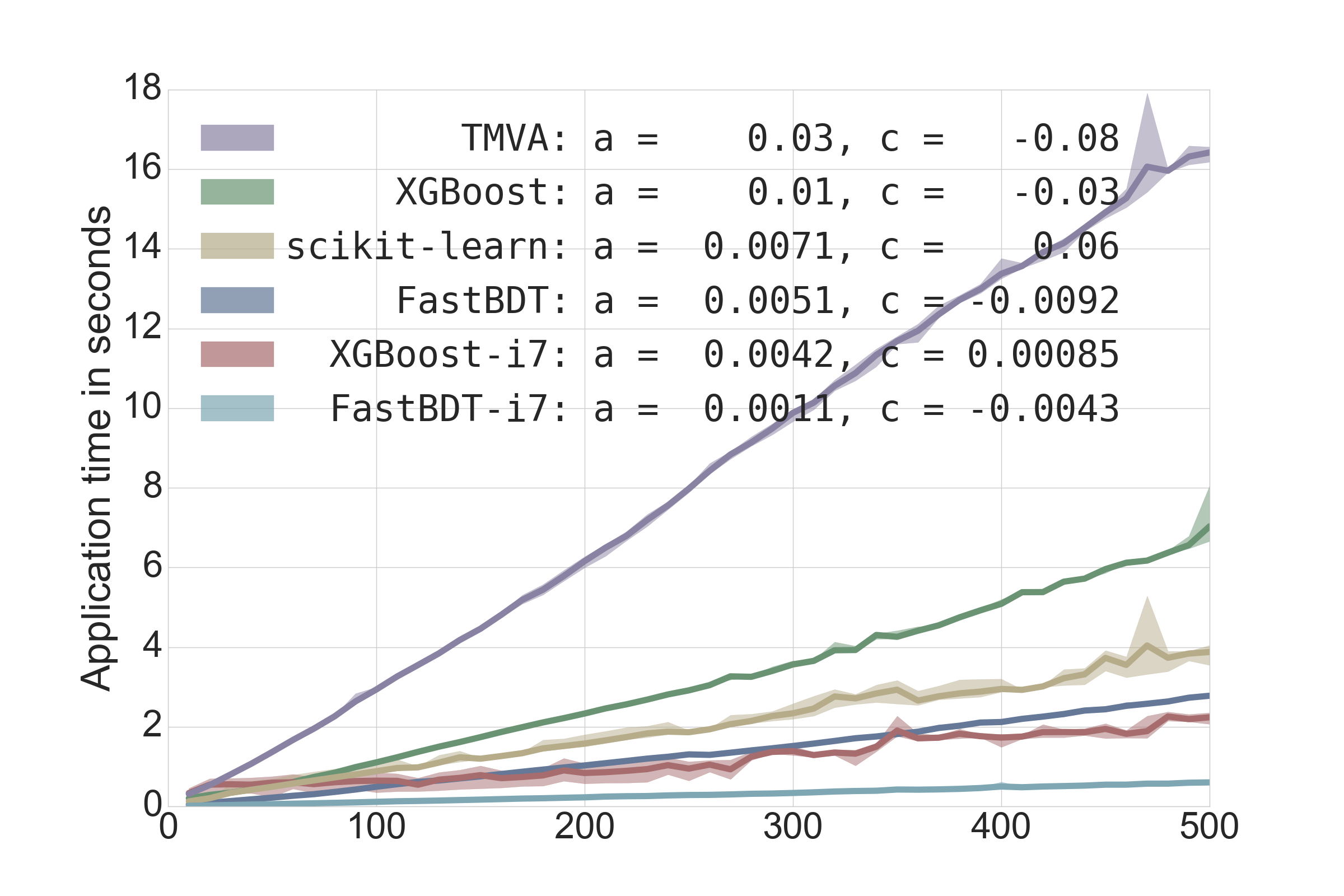}
\caption{\texttt{number of trees}}
\label{fig:ApplynTrees}
\end{subfigure}\hfill
\begin{subfigure}[b]{0.49\textwidth}
\includegraphics[width=1.1\textwidth]{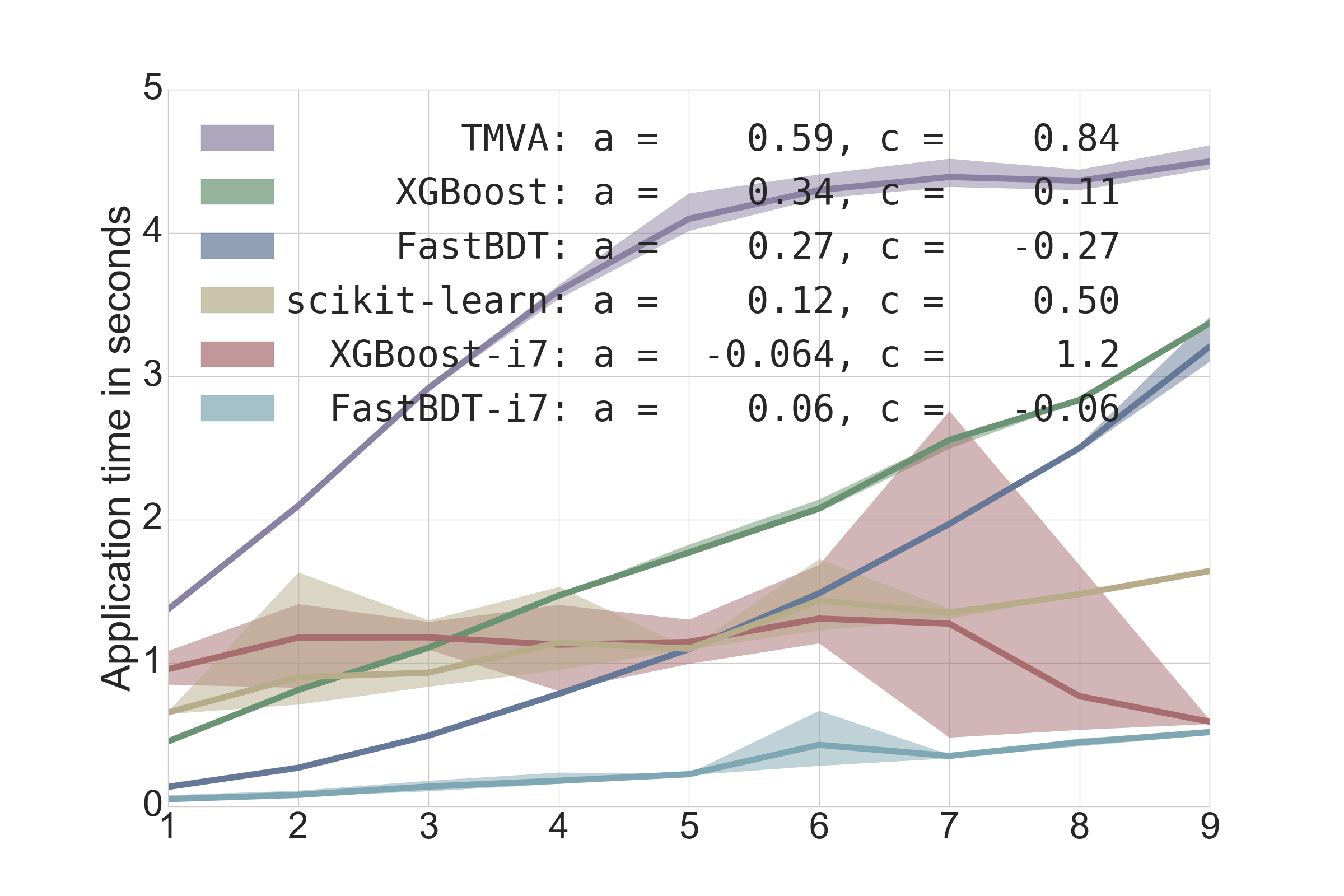}
\caption{\texttt{depth of the trees}}
\label{fig:Applydepth}
\end{subfigure}\hfill
\begin{subfigure}[b]{0.49\textwidth}
\includegraphics[width=1.1\textwidth]{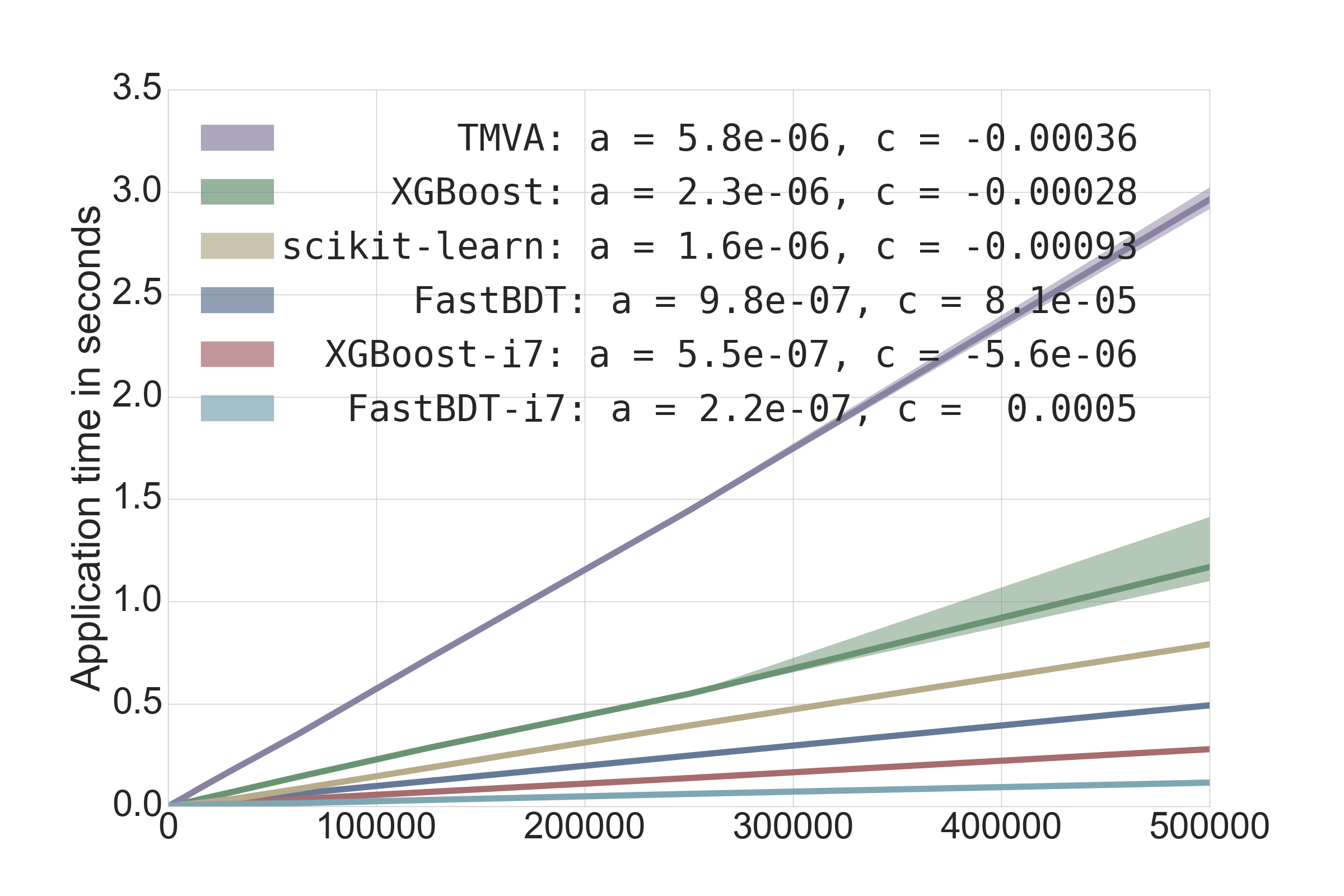}
\caption{\texttt{number of test data-points}}
\label{fig:ApplynEvents}
\end{subfigure}\hfill
\begin{subfigure}[b]{0.49\textwidth}
\includegraphics[width=1.1\textwidth]{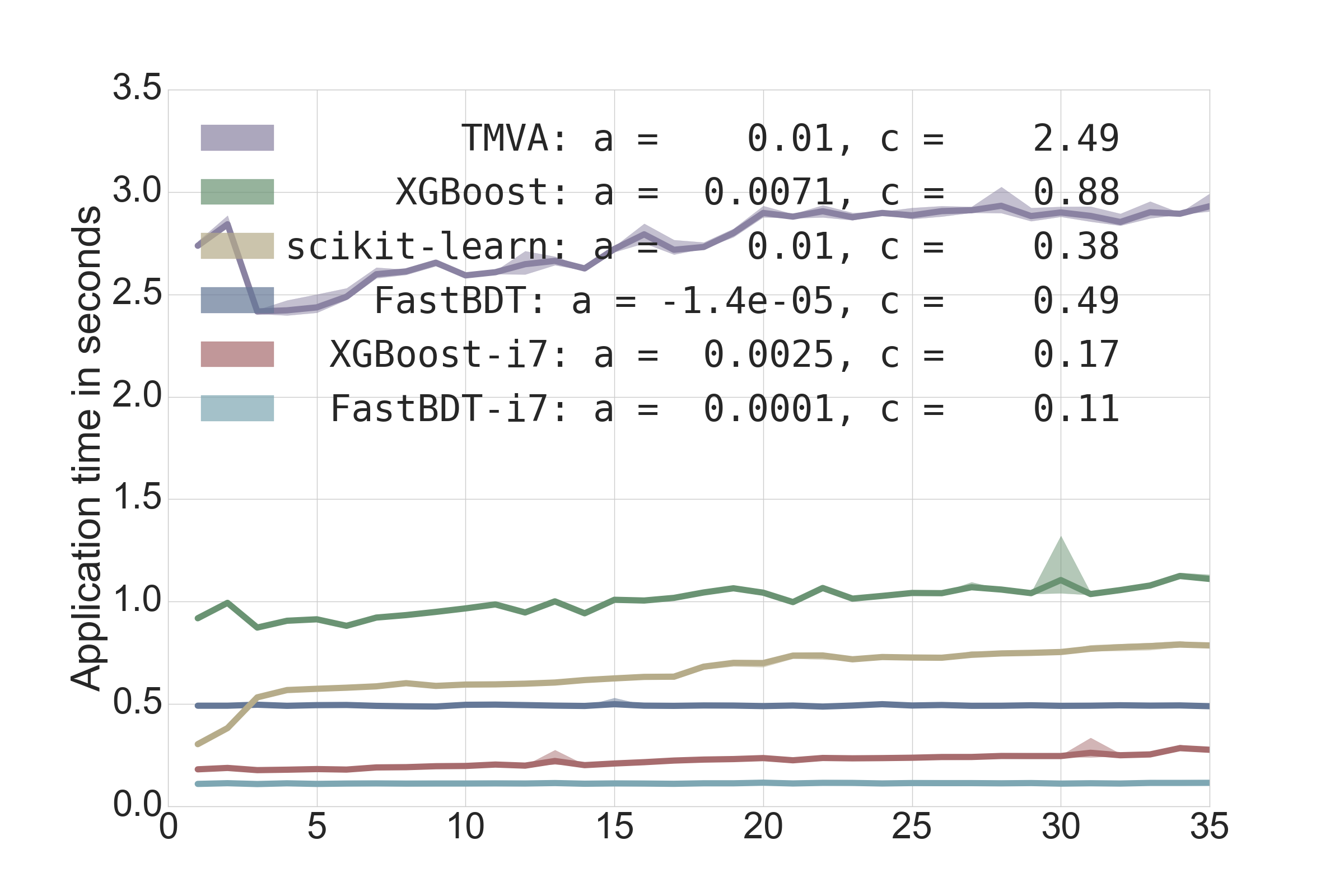}
\caption{\texttt{number of features}}
\label{fig:ApplynFeatures}
\end{subfigure}
\begin{subfigure}[b]{0.49\textwidth}
\includegraphics[width=1.1\textwidth]{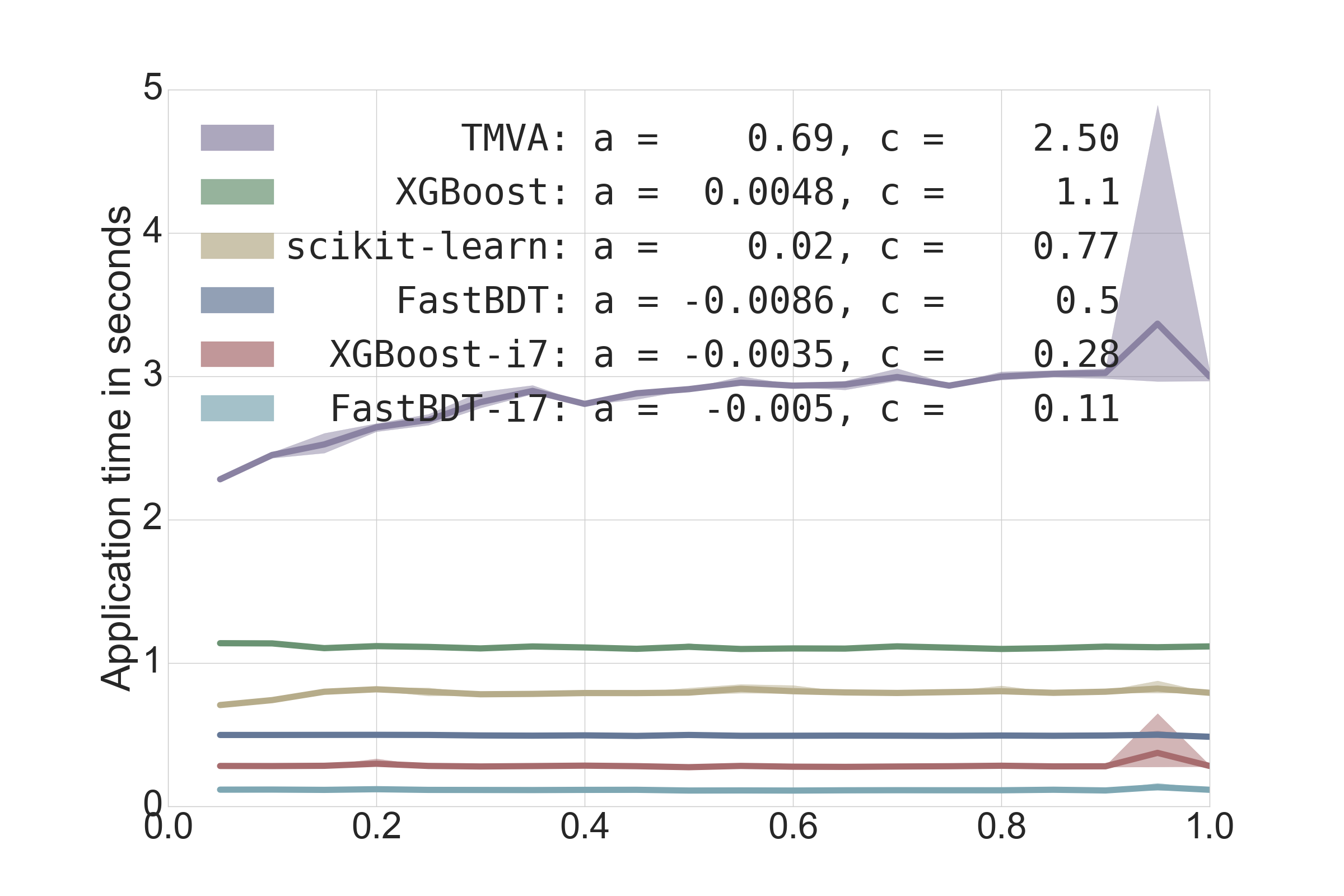}
\caption{\texttt{sampling rate}}
\label{fig:ApplySamplingRate}
\end{subfigure}
\caption{Runtime behaviour during the appplication-phase of different hyper-parameters. The $y$ axes show the application time. The $x$ axes show the varied hyper-parameter. Each configuration was measured five times:
The mean values are shown as solid lines, whereas the minimal (maximal) values define the lower (upper) bound of the shaded band around the lines. A linear regression $y = a \cdot x + c$ was performed
using the minimal values of each configuration. The order of the legend entries is the same as the position of the solid lines at the right limit of each plot.}
\label{fig:TestTime}
\end{figure}

During the application-phase the runtime should scale linearly in  \texttt{depth of the trees}, \texttt{number of trees} and
\texttt{number of test data-points}. The runtime should be independent of the \texttt{sampling rate} and \texttt{number of features}.
The results are summarized in Figure \ref{fig:TestTime}. Detailed speedup comparisons between all tested implementations are stated in Table \ref{tab:applicationphase}.

The single-core version of FastBDT outperforms all other single-core implementations during the fitting phase.
The multi-core version of FastBDT is on average $3.8$ times faster than the multi-core version of XGBoost.
The runtime behaviour in the \texttt{number of trees} and \texttt{number of test data-points} is as expected.
TMVA seems to be faster for small values of the \texttt{sampling rate} and again shows a constant runtime
behaviour for \texttt{depth of the trees} $> 6$. Both can be explained by the limit on the minimum number of data-points per node.

FastBDT violates the expected linear scaling in the \texttt{depth of the trees}. Probably the caching does not work optimally
for deep trees.

\begin{table}
\centering
\setlength\tabcolsep{2pt}
\caption{Average ratio of application-phase runtime over all considered hyper-parameter configurations.}
\label{tab:applicationphase}
\begin{tabular}{c|rrrrrr}
	\toprule
	           & FastBDT & FastBDT-i7 & scikit-learn & XGBoost & XGBoost-i7 &  TMVA \\ \midrule
	 FastBDT   &         &   $0.3$    & $1.5$  & $2.2$  & $0.9$  & $5.7$   \\
	FastBDT-i7 &   $4.3$ &            & $6.4$  & $9.4$  & $3.8$  & $24.3$  \\
	 scikit-learn   &   $0.7$ &   $0.2$    &        & $1.5$  & $0.6$  & $3.9$   \\
	 XGBoost   &   $0.5$ &   $0.1$    & $0.7$  &        & $0.4$  & $2.6$   \\
	XGBoost-i7 &   $1.6$ &   $0.4$    & $2.2$  & $3.4$  &        & $8.9$   \\
	   TMVA    &   $0.2$ &   $0.1$    & $0.3$  & $0.4$  & $0.2$  &         \\ \bottomrule
\end{tabular}
\end{table}

\subsection{Classification Quality}
The quality of the classification was evaluated using the area under the receiver operating characteristic (ROC) curve.
This quality indicator is independent of the chosen working-point (i.e., the desired efficiency or purity) and is widely
used to compare the separation power of different algorithms to one another.
The higher the value the better the separation between signal and background.

The main difference between the implementation arises due to the different regularisation methods.
FastBDT uses the equal-frequency binning to prevent over-fitted cuts; XGBoost
employs a modified separation gain, which includes the structure of the current tree;
and TMVA limits the minimum number of data-points per event to $5 \%$.

FastBDT outperforms the other implementations in most situations, except for extremely deep trees.
In this region the over-fitting effect degrades the performance of the classifier, whereas XGBoost
seems to have a superior regularisation method to prevent over-fitting in this situation.
The performance of FastBDT and XGBoost is nearly independent of the used sampling rate,
whereas TMVA and scikit-learn strongly depend on it. Hence it is likely
that TMVA and scikit-learn over-fit in regions with small sampling rate (and therefore low statistic).

The classification quality will be different for other datasets, and depends strongly on
the concrete classification problem at hand.

\begin{figure}
\centering
\begin{subfigure}[b]{0.49\textwidth}
\includegraphics[width=1.1\textwidth]{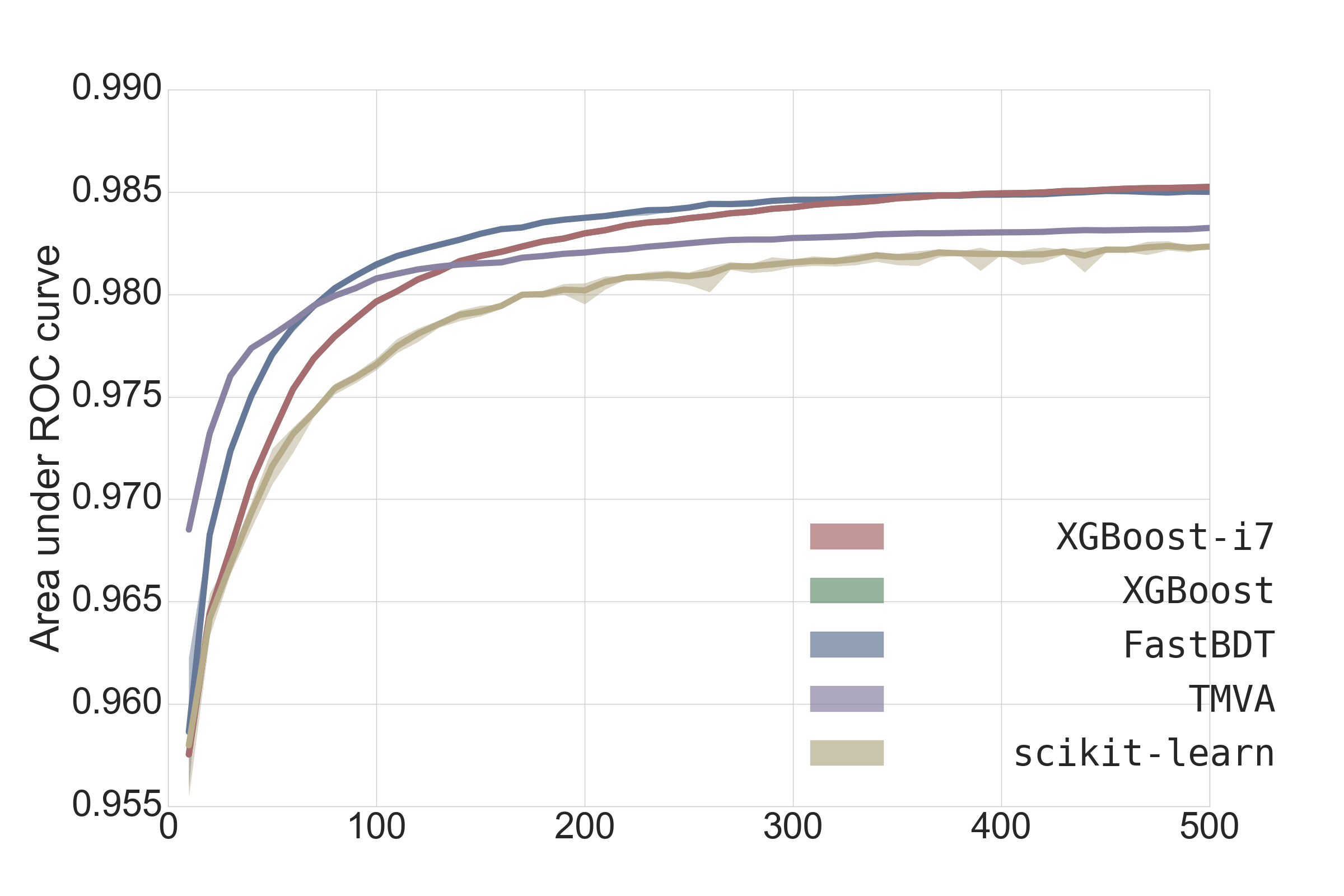}
\caption{\texttt{number of trees}}
\label{fig:AUCnTrees}
\end{subfigure}\hfill
\begin{subfigure}[b]{0.49\textwidth}
\includegraphics[width=1.1\textwidth]{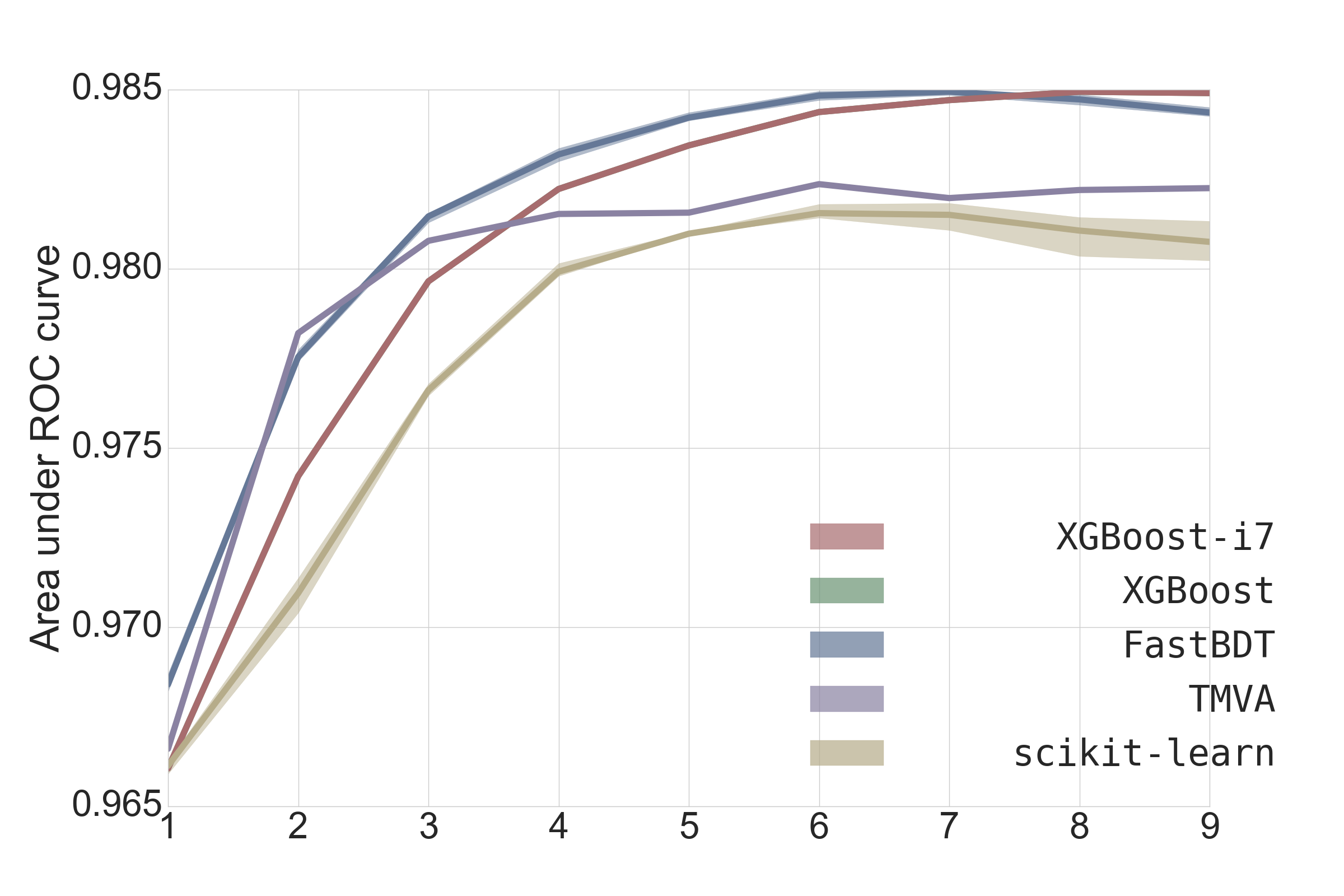}
\caption{\texttt{depth of the trees}}
\label{fig:AUCdepth}
\end{subfigure}\hfill
\begin{subfigure}[b]{0.49\textwidth}
\includegraphics[width=1.1\textwidth]{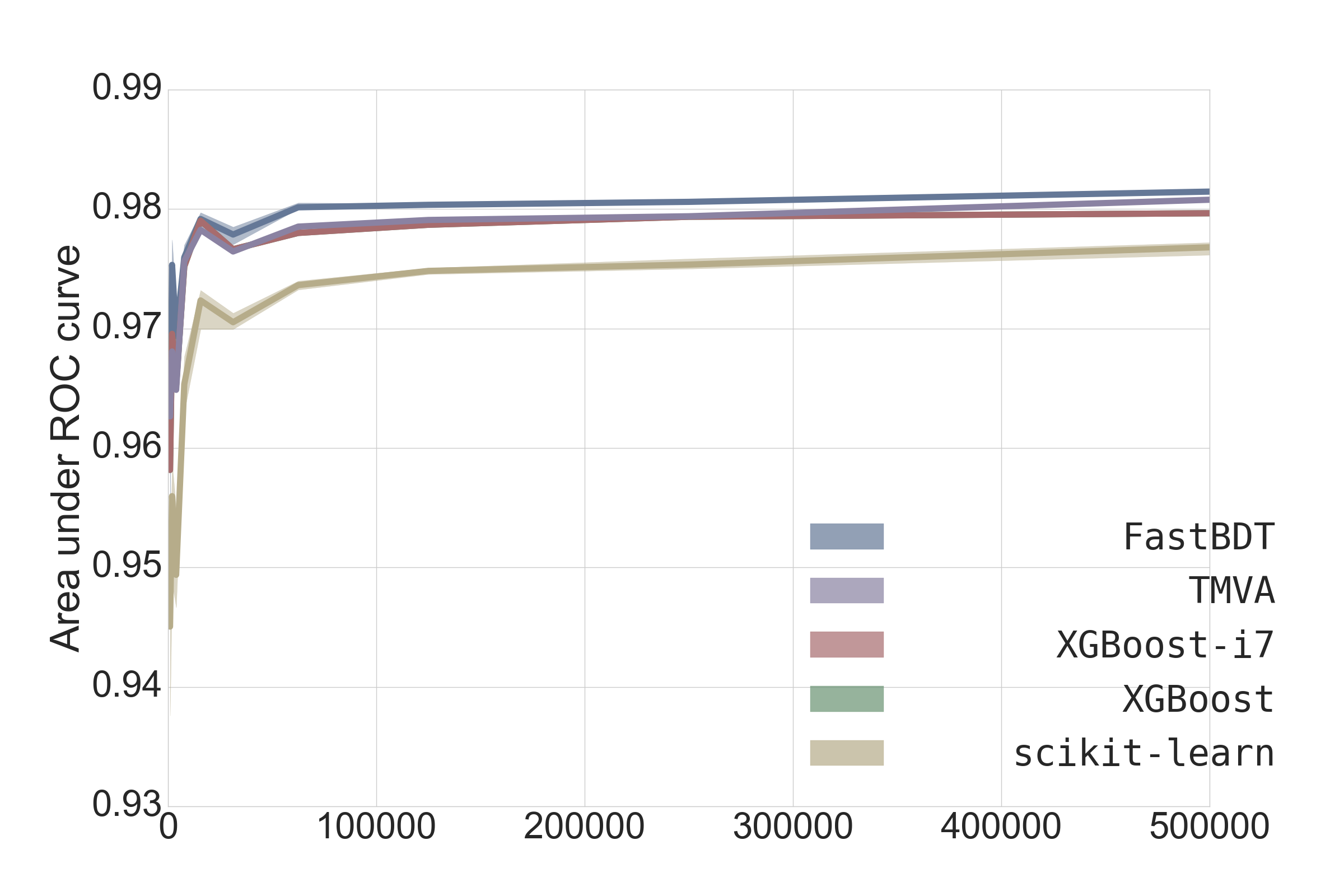}
\caption{\texttt{number of training data-points}}
\label{fig:AUCnEvents}
\end{subfigure}\hfill
\begin{subfigure}[b]{0.49\textwidth}
\includegraphics[width=1.1\textwidth]{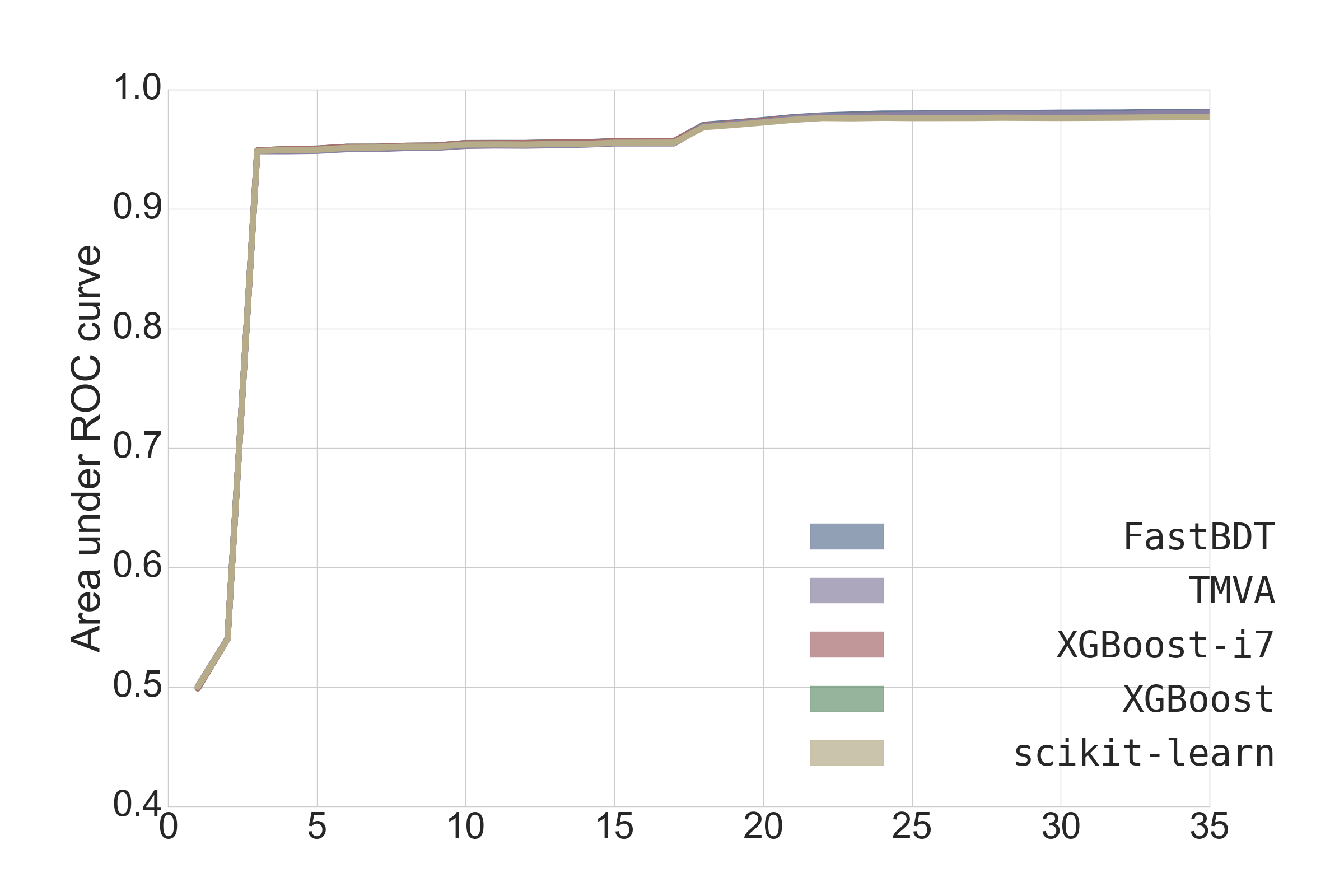}
\caption{\texttt{number of features}}
\label{fig:AUCnFeatures}
\end{subfigure}
\begin{subfigure}[b]{0.49\textwidth}
\includegraphics[width=1.1\textwidth]{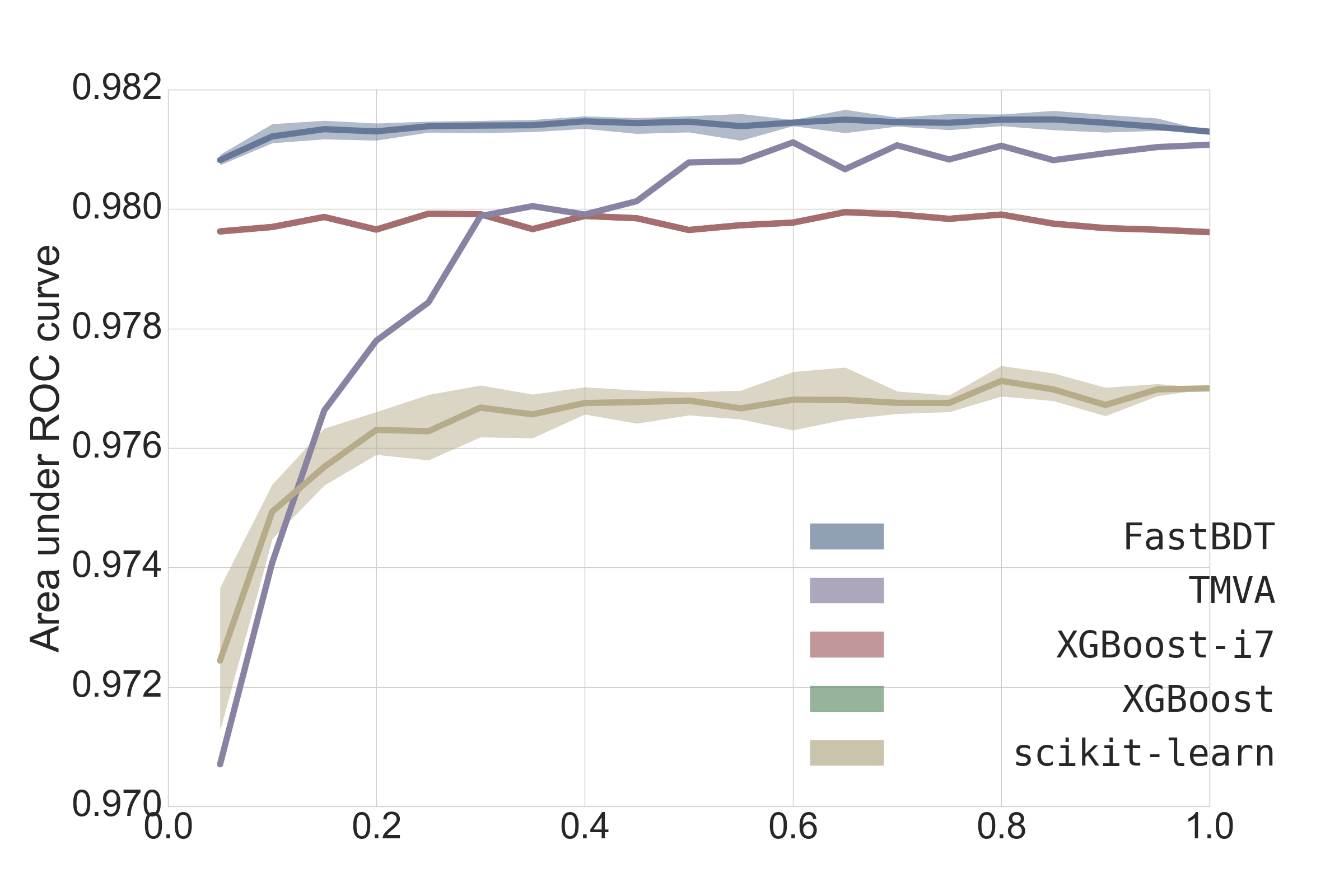}
\caption{\texttt{sampling rate}}
\label{fig:AUCnSamplingRate}
\end{subfigure}
\caption{Behaviour of the classification quality with different hyper-parameter configurations. The $y$ axes show the area under the receiver operating characteristic (ROC) curve. The $x$ axes show the varied hyper-parameter. Each configuration was measured five times:
The mean values are shown as solid lines, whereas the minimal (maximal) values define the lower (upper) bound of the shaded band around the lines. The order of the legend entries is the same as the position of the solid lines at the right limit of each plot. Both XGBoost versions yield identical results, therefore the curves are on top of each other and only one can be seen.}
\label{fig:AUC}
\end{figure}

\section{Advanced Features}
FastBDT offers more advanced capabilities, three of them are briefly described in the following.

\subsection{Support for Negative Weights}
The boosting algorithm assigns a weight to each data-point in the training dataset.
FastBDT supports an additional weight per data-point provided by the user.
These individual weights are processed separately from the boosting weights.
In particular FastBDT allows for negative individual weights, which are commonly used in data-driven techniques
to statistically separate signal and background using a discriminating variable \citep{Martschei:2012pr}.
Other frameworks like TMVA, SKLearn and XGBoost support negative weights as well.

\subsection{Support for Missing Values}
There are at least two different kinds of missing values in a dataset.
Firstly, missing values which can carry usable information about the target e.g. in particle classification in HEP a feature provided by a detector can be absent because the detector was not activated by the particle.
Secondly, missing values which should not be used to infer the target e.g. a feature provided by a detector is absent due to technical
reasons.

FastBDT supports both types of missing values.
The first type can be passed as negative or positive infinity, FastBDT will put these values in its underflow or overflow bin.
In consequence, a cut can be applied separating the missing from the finite values, and the method can use the information provided
by the presence of a missing value.
The second type should be passed as NaN (Not a Number) floating point value according to the IEEE 754 floating point standard \citep{ieee754}.
These values are ignored during the fitting and application phase of FastBDT. If the tree tries to cut on a feature, which is NaN,
the current node behaves as if it is a terminal-node for the corresponding data-point.

\subsection{Feature Importance Estimation}
May multivariate classification methods offer the possibility to estimate the feature importances, meaning influence
of the features on the decision. For BDTs the usual approach to calculate the global feature importance,
is to sum up the separation gain of each feature, by looping over all trees and nodes.
The individual importance for a single data-point can be calculated similarly by summing up all separation gains
along the path of the event through the trees.

This approach suffers from the possible correlation and non-linear dependencies between the features,
as can be seen by the simple example in Table \ref{table:importance}. In a single decision tree, one of the features
will have a separation gain of zero, although both carry exactly the same amount of information.

\begin{table}
\centering
\caption{Truth table of exclusive or. Both input features $x$ and $y$ cannot classify the target on their own.}
\begin{tabular}{ccc}
\toprule
$x$ & $y$ & Target \\ \midrule
1 & 1 & 0 \\
1 & 0 & 1 \\
0 & 1 & 1 \\
0 & 0 & 0 \\ \bottomrule
\end{tabular}
\label{table:importance}
\end{table}

Due to the fast fitting phase of FastBDT it is viable to use another popular approach, by measuring the decrease
in the performance (e.g. the area under the curve (AUC) of the receiver operating characteristic) of the method if a feature is left out.
This requires $N$ fit operations, where $N$ is the number of features.
We can improve the accuracy of the method further by recursively eliminating the most-important feature, 
which requires $\frac{1}{2} N ( N + 1) $ fit operations.

Although this approach is independent of the employed classification method, it only becomes a viable option
with a fast and robust method.

\section{Conclusion}
FastBDT implements the widely employed Stochastic Gradient-Boosted Decision Tree (SGBDT) algorithm \citep{Friedman2002367}, which exhibits a good
out-of-the-box performance and generates an interpretable model. Often, multivariate classifiers are trained once
and are applied on big data-sets afterwards. FastBDT performs well in this use-case and offers support for missing values,
an equal-frequency preprocessing of the features and a fast application-phase.
The main advantage is the fast (in terms of CPU time) fitting-phase compared with popular implementations like TMVA \citep{Hocker:2007ht}, scikit-learn \citep{scikit-learn} and XGBoost \citep{XGBoost}.
Possible use-cases are: Real-time learning applications, frequent re-fitting of classifiers, fitting a large number of classifiers, and measuring the dependence of many variables
to one another.

FastBDT outperforms other popular implementations in terms of runtime during the fitting-phase and application-phase,
as well as in the final classifier quality. The presented runtime measurements are independent of the data-set.
The techniques employed by FastBDT could be migrated to other implementations, especially the equal-frequency binning works well
with boosted-decision trees.

FastBDT is licensed under the GPLv3 license and available on github \url{https://github.com/thomaskeck/FastBDT}.

\bibliography{bibtex}

\end{document}